\documentclass{elsarticle}

\usepackage{lineno,hyperref}
\usepackage{amssymb,amsmath,array}
\usepackage{adjustbox}
\usepackage{makecell}
\usepackage{verbatim}
\usepackage{chngpage}
\usepackage{booktabs}

\modulolinenumbers[5]

\journal{Neurocomputing}

\begin{document}

\begin{frontmatter}

%\title{Graph generation by sequential edge prediction} %OLD
%\title{Graph generation by recurrent neural networks edge prediction} %Alessio
\title{Edge-based sequential graph generation with recurrent neural networks} %Marco

% \tnotetext[mytitlenote]{Fully documented templates are available in the elsarticle package on \href{http://www.ctan.org/tex-archive/macros/latex/contrib/elsarticle}{CTAN}.}

%% Group authors per affiliation:
\author[unipi]{Davide Bacciu}
\ead{bacciu@di.unipi.it}
\author[unipi]{Alessio Micheli}
\ead{micheli@di.unipi.it}
\author[unipi]{Marco Podda}
\ead{marco.podda@di.unipi.it}

\address[unipi]{Dipartimento di Informatica,
Universit\`a di Pisa,
Italy.}

\begin{abstract}
Graph generation with Machine Learning is an open problem with applications in various research fields. In this work, we propose to cast the generative process of a graph into a sequential one, relying on a node ordering procedure. We use this sequential process to design a novel generative model composed of two recurrent neural networks that learn to predict the edges of graphs: the first network generates one endpoint of each edge, while the second network generates the other endpoint conditioned on the state of the first. We test our approach extensively on five different datasets, comparing with two well-known baselines coming from graph literature, and two recurrent approaches, one of which holds state of the art performances. Evaluation is conducted considering quantitative and qualitative characteristics of the generated samples. Results show that our approach is able to yield novel, and unique graphs originating from very different distributions, while retaining structural properties very similar to those in the training sample. Under the proposed evaluation framework, our approach is able to reach performances comparable to the current state of the art on the graph generation task.
\end{abstract}

\begin{keyword}graph generation; recurrent neural networks; auto-regressive models; deep learning
\end{keyword}

\end{frontmatter}

% \linenumbers

\section{Introduction}\label{sec:introduction}
Graphs are well-known data structures that allow to store and access relational data efficiently. Their use to represent information is ubiquitous, especially in domains such as Biology \cite{huber01}, Chemistry \cite{estrada02} and Natural Language Processing \cite{nastase03}. In all these fields, as well as many others, data do not exist in isolation, but are connected among themselves by complex relationships. Hence, graphs are usually preferred to "flat" vectorial data whenever there is the need to encode both relational knowledge and numerical information in a concise and compact way. Given the superior expressiveness of graphs, many Machine Learning models have been developed to process them as input samples, with the aim of creating more informed predictors on a variety of tasks. This trend has been increasing especially since the advent of Graph Neural Networks \cite{scarselli04} and contextual Neural Networks for Graphs \cite{micheli09}, which paved the road for modern graph-based Deep Learning \cite{bronstein05} models. As of today, Graph Neural Networks are used with success for predictive tasks such as semi-supervised classification \cite{kipf06}, link prediction \cite{zhang07}, and text classification \cite{yao08}.

Besides being able to predict outcomes using graphs, one open and less studied problem in Machine Learning is how to instruct learning models to generate graphs from arbitrary distributions.
This task is indeed hard for a number of reasons:
\begin{itemize}
    \item the space of graphs is exponential: for a fixed number of nodes $N$ there are $2^{N(N-1)/2}$ possible undirected graphs. This implies that to learn a graph distribution, one cannot aim to explore the entire graph space, except for trivial instances. Moreover, graph distributions of interest usually cover only a tiny portion of this large space. Hence, to be successful, generative models need to focus on small subspaces of high probability, while avoiding regions of low or zero probability.
    \item graphs are discrete objects by nature. This means that standard machine learning techniques, which were developed primarily for continuous, vectorial data, do not work off-the-shelf, but usually need adjustments. A prominent example is the back-propagation algorithm, which is not directly applicable to graphs, since it works only for continuously differentiable objective functions. Indeed, the usual way to deal with this issue is to project graphs (or their constituents) into a continuous space and represent them as vectors, with a process known as graph embedding \cite{paassen39, hamilton09}.
    \item real-world graph are generated according to very complex structural relationships. This complicates the process of developing useful graph likelihood functions which can then be optimized by generative models (as in \cite{cgmm}), since one cannot usually rely on simplifications such as, for example, assuming independence between edges.
\end{itemize}
In this work, we present a generative model for graphs that tries to address these challenges. 

To deal with the first issue, we propose to move the generative process from the complex space of graphs to the more "friendly" space of sequences, casting the generative process into a sequence of nodes and edges predictions. Note that generating a graph by adding one node or edge at a time most likely resembles the way a human would think about graph generation (see e.g. \cite{cogmaps}). Besides being intuitive, this choice has the positive effect of shrinking the search space of interesting graphs, because we are now restricting ourselves to graphs that are formed \emph{in a particular order}.

As regards the second and third issues, representing the generative process of graphs in this way allows us to learn it sequentially. This, in turn, means that reliable learning models such as Recurrent Neural Networks (RNNs) \cite{elman10}, which are more easily optimizable, can now be used. Moreover, we have now a way to learn dependencies without having to resort to simplifying assumptions that can prevent a model from learning effectively.

Summing up, in this work we present the following contributions:

\begin{itemize}
    \item we introduce a novel general-purpose generative model for graphs. First, as a preprocessing step, we order graph nodes according to a breadth-first visit of the graph. Then, we formulate the generative process as a sequence of two auto-regressive tasks, which can be used to infer the sets of nodes and edges of training graphs, as well as to generate new ones;
    \item we experiment extensively with the proposed model, comparing its performances with strong baselines, as well as state-of-the-art, Deep-Learning based approaches. Our experimental framework has been designed to evaluate the proposed model on multiple levels, including a quantitative aspect related to the ability of the model to produce novel and diverse graphs, and a qualitative aspect to assess whether the graphs generated by the model retain the structural features of the ones in the training sample;
    \item we show that, under our evaluation framework, the proposed model is able to perform at, and sometimes surpass, the state-of-the-art in the task of generating graphs coming from very different distributions, outperforming baseline models, sometimes by a very large margin;
    \item we study the effect of various node ordering functions, by comparing our approach with variants trained using different node orderings, showing that the adopted node ordering strategy is more effective for learning complex dependencies than the alternatives, and produces graphs of better quality.
\end{itemize}
This work is structured as follows: Section \ref{sec:related_work} provides a gentle introduction to the subject, describing historical as well as present approaches to graph generation, and highlighting key differences among current generative models and ours. In Section \ref{sec:notation}, we introduce the necessary mathematical notation used to formulate the model, which is then expounded in Section \ref{sec:model}. In particular, we describe its training and inference procedures in depth, and provide details on the implementation. In Section \ref{sec:experiments}, we disclose the experiments, describing the datasets that were used, the baselines against which we compared, as well as the evaluation framework used to assess the validity of our approach. In Section \ref{sec:results}, we present and discuss the results of the experiments. In Section \ref{sec:ordering}, we study the effect of various node ordering functions on the performances of the model. Finally, in Section \ref{sec:conclusions}, we draw our conclusions and highlight future research directions.

\section{Related Work}\label{sec:related_work}
Graph generation is an interesting problem that has been tackled from several viewpoints. The first generative models of graphs were conceived by graph theorists, with the aim of studying certain graph properties of interest, such as connectivity behavior in relation to graph size. This research branch is well represented by three renowned models: the Erd\"{o}s-R\'{e}nyi (ER) model \cite{erdos11}, the Barab{\'a}si-Albert (BA) model \cite{barabasi12}, and the Watts-Strogatz (WS)  model \cite{watts14}. The ER model is the first generative model in history. While useful for theoretical purposes, its main limitations is that it assumes edge independence, thus being unsuitable to approximate real-world graph distributions that display strong edge dependencies. The BA model is thought for social networks, and allows to generate graphs whose node degree follows a scale-free distribution, while the WS model can generate graphs with the so-called "small-world property", i.e. graphs with short average path length and high clustering coefficients. However, both are able to model only their specific property, and fail at generalizing to more diverse graphs: for example, the BA model struggles at reproducing graphs with the small-world property, and vice versa, the WS model poorly approximates scale-free node degree distributions. One limitation that all these three models have in common is that their parameters cannot be learned from data, but must be fixed beforehand.

As such, especially with the late explosion of deep generative learning, an increasing number of methods to learn graph distributions adaptively from data have been designed. Initially, generative models have been mostly used to capture the distribution of structured samples, such as trees \cite{bhtmm} or graphs \cite{cgmm}, for the sole purpose of constructing predictors (i.e. classifiers or regressors) for structured data \cite{genkerntreestructureddata}. Lately, attention is shifting towards the generative use of such learned distribution. For example, Deep Learning models have been used with success to generate parse trees \cite{deepTreeTransd,socher13}, intermediate representations of reasoning tasks \cite{johnson15}, and more broadly for general structure-to-structure transduction \cite{sun16}. One domain where deep generative models for graphs are used extensively is Cheminformatics. Indeed, since molecules can naturally be thought as graphs, models able to generate such structures are widely adopted and increasingly developed for tasks such as drug discovery, drug property optimization, and protein structure generation \cite{namrata20}. In this domain, two broad classes of generative models for graphs have emerged. The first one is based on encoder-decoder architectures: the encoder maps the input molecular graph to a Gaussian latent space, whose sampling parameters are learned with a Variational Auto-Encoder \cite{vae}; typical decoders either reconstruct the adjacency matrix with a pairwise dot-product based neural network \cite{kipf22, grover19}, or transform molecules into SMILES \cite{weininger37} strings and learn the corresponding language model using RNNs \cite{gomez-bombarelli23, kusner24}. The other class of generative models stems from the seminal work in \cite{johnson15}, which defines a framework for taking a graph as input and apply some transformations (such as node and edge addition and state update) to produce a different graph as output. Notable members of this family of molecule generators are presented in \cite{liu17, jin18}.

Works such as in \cite{gomez-bombarelli23, kusner24} share with ours the use of RNNs (especially as decoders). Indeed, RNNs are universal function approximators for sequence processing.
The core concept of RNNs is the notion of recurrent state, which is the internal ”memory”
used to keep information about previous elements in the sequence. Such state is updated at each step according to the currently processed sequence element and previous state values. We employ RNNs as the basic building blocks of our model, since we aim to generate graphs in a sequential fashion.

In contrast with the above-mentioned works, rather than focusing on a specific domain, we seek to develop a general purpose generative model, that is able to learn arbitrary graph distributions without making strong assumptions on the underlying nature of the graphs of interest. With this distinction in mind, two models that share our goal are the ones proposed in \cite{li33} and \cite{you34}. With both models, we share the choice of representing the generative process of a graph as a sequence.

The model in \cite{li33} frames the generation of a graph as a decision-making process, where nodes and edges are generated according to a sequence of binary decisions, learned adaptively from data using Multi-Layer Perceptrons, whose output is used to condition the state of the current graph being generated. More precisely, at each step the model takes the decision of whether to add a new node; if added, it checks whether to add new edges for the node; if so, it iteratively adds all edges of that node to the current graph. The generated graph state is maintained and updated using the typical message-passing propagation scheme of Graph Neural Networks \cite{scarselli04}. This model is expressive and proven to generalize to very diverse graph distributions; however, the complexity of the generative process is $O(MN^2diam(G))$, where $diam$ is the diameter of the graph being generated \cite{you34}. One clear difference between our proposal and this approach is that we do not rely on a message-passing scheme to maintain the state of the graph being generated, but we rather use RNNs.

The model in \cite{you34}, also termed GraphRNN (GRNN), treats the generation of a graph as an auto-regressive process, modeled with Recurrent Neural Networks. More precisely, at a given step of the computation, each newly added node generates a binary sequence representing the adjacency between itself and all previously generated nodes. This generation is conditioned on an encoding of the graph being generated, which is also maintained and updated using a different RNN. To avoid long-term dependencies issues when learning from graphs with a large number of nodes, the authors propose a method to constrain the length of the binary sequence vector by exploiting a breadth-first ordering of the nodes. As such, time complexity for graph generation is improved to $O(mN)$, where $m$ is the maximum length of the constrained binary sequence vector. This model holds state-of-the-art performances as regards the quality of generated graphs. One key difference between this model and our approach is discussed in Section \ref{sec:preprocessing}; furthermore, we do not predict adjacencies using binary vectors, but act on the distribution of nodes directly.

\section{Notation and Definitions}\label{sec:notation}
A graph is a pair $G=\langle V, E \rangle$, where $V = \{v_0, v_1, \ldots, v_N\}$ is a set of $N$ nodes (or vertices) and $E = \{ (v, u) \mid v, u \in V\}$ is a set of $M$ edges (or links). Let us assume without loss of generality that the graphs we deal with are fully connected (i.e. they are formed by a single connected component), and do not contain self-loops (i.e. edges of the form $(v, v)$). Given a generic edge $(v, u)$, for ease of notation we will call $v$, the first node in the edge pair, the \emph{source} node, and $u$, the second node in the edge pair, the \emph{destination} node. We furthermore assume the existence of a bijective labeling function $\pi: V \rightarrow \mathbb{N}_0$ that assigns a non-negative integer (a \emph{node ID}) to each node in the graph according to some ordering criteria. If $\pi$ is given, we can represent $G$ with its \emph{ordered edge sequence} $s$, defined as:
$$s = (s_1, \ldots, s_M),\; \mathrm{with}\; s_i = (x_i, y_i),\; \mathrm{where}\; (\pi^{-1}(x_i), \pi^{-1}(y_i)) \in E.$$ 
Note that the sequence is ordered, that is we impose $s_i \leq s_{i+1}\; \mathrm{iff}\; x_i < x_{i+1}\; \mathrm{or}\; (x_i = x_{i+1}\; \mathrm{and}\; y_i < y_{i+1})$. In other terms, $s$ is sorted in ascending lexicographical order (note that such lexicographical order is unique, since we are excluding self-loops). Elements of $s_i$ are pairs of integers $(x_i, y_i)$ assigned to nodes by $\pi$; thus, every sequence element $s_i$ identifies an edge by specifying the ID of nodes that share it. If an ordered edge sequence is given, the sets $V$ and $E$ of $G$ can be uniquely reconstructed. Lastly, let us define functions $\mathcal{X}(s) = (x_1, \ldots, x_{M})$ and $\mathcal{Y}(s) = (y_1, \ldots, y_{M})$, which extract source and destination nodes, respectively, from an ordered edge sequence $s = ((x_1, y_1), \ldots, (x_M, y_M))$, keeping the same order. We will refer to $\mathcal{X}$ and $\mathcal{Y}$ as source and destination sequences, dropping the dependency of $\mathcal{X}$ and $\mathcal{Y}$ on $s$ for ease of notation. As a final remark, note that the ordered edge sequence of a graph is dependent on the particular choice of $\pi$; that is, different node ordering strategies lead to different edge sequences. We discuss later in Section \label{sec:model} how $\pi$ can be chosen and assess the impact of this choice on the structures generated by our model in Section \ref{sec:ordering}.

\section{Model}\label{sec:model}
Learning a graph distribution of interest amounts to estimating a probability distribution $P(\mathcal{G})$ from a finite sample of structures $\mathcal{G}$. New graphs can then be generated from $P(\mathcal{G})$ by an appropriate sampling process. Both learning and sampling directly from $P(\mathcal{G})$ are hard except for trivial distributions, as discussed in Section \ref{sec:introduction}. To overcome this limitation, in this work, we learn an alternative task: we first represent graphs with their associated ordered edge sequences, and then maximize their probability:

$$\max P(S_{\mathcal{G}}) = \prod_{s \in S_\mathcal{G}} P(s),$$
where $S_{\mathcal{G}}$ is a dataset of ordered edge sequences extracted from the corresponding graph training sample, $\mathcal{G}$.
To learn this  probability, we represent the generative process of an ordered edge sequence as follows:
$$ 
P(s) = P(\mathbf{Y}\, |\, \mathbf{X})\; P(\mathbf{X}),
$$
where $\mathbf{X}$ and $\mathbf{Y}$ are random variables associated to $\mathcal{X}$ and $\mathcal{Y}$, respectively. Informally, a graph ordered sequence can be generated by first sampling its source nodes sequence, and then using it to sample the destination sequences. 

This generative process has an intuitive interpretation in the domain of graphs: $P(\mathbf{X})$ acts as a "soft prior" on the structure of the graph: for example, it implicitly specifies the minimum outward degree of some nodes (which can be thought as the "main" nodes driving the generation), represented as the number of times the node ID appears in the sequence. In contrast, $P(\mathbf{Y}|\mathbf{X})$ explicitly models the connectivity of the graph conditioned on such prior information, assigning elements of $\mathcal{X}$ their adjacent nodes. From a different angle, it is responsible of improving (or leave unchanged) the minimum degree imposed by the prior $P(\mathbf{X})$.

We choose to approximate $P(\mathbf{X})$ and $P(\mathbf{Y}|\mathbf{X})$ with deep auto-regressive neural networks. More in detail, $P(\mathbf{X})$ is implemented as an auto-regressive model that approximates the prior as $p(x_{i+1} \mid x_{i}, \mathbf{h}^{p}_{i})$, $i \in 1, \ldots, M$, and $P(\mathbf{Y}|\mathbf{X})$ is implemented as another auto-regressive model that approximates the conditional with a function $q(y_i \mid x_i,\,\mathbf{h}^{q}_{i})$. In both models, the $\mathbf{h}$ vector is the hidden recurrent state of the sequence at time step $i$. Following, we will sometimes refer to these two networks as RNN1 and RNN2, respectively. In our implementation, the hidden recurrent state of RNN2 is initialized using the last hidden recurrent state of RNN1. Note that the recurrent states of the two networks have themselves a nice interpretation in the domain of graphs: for RNN1, the state can be thought of as a running encoding of the soft prior structure used to condition the generative process; for RNN2, it can be thought of as a graph embedding that gets updated as new nodes are connected to the current graph. 
Following, we describe how such a model is handled during training and inference phases.

\subsection{Preprocessing and node ordering}\label{sec:preprocessing}
\begin{figure}[h!]
\centering
\includegraphics[width=\textwidth]{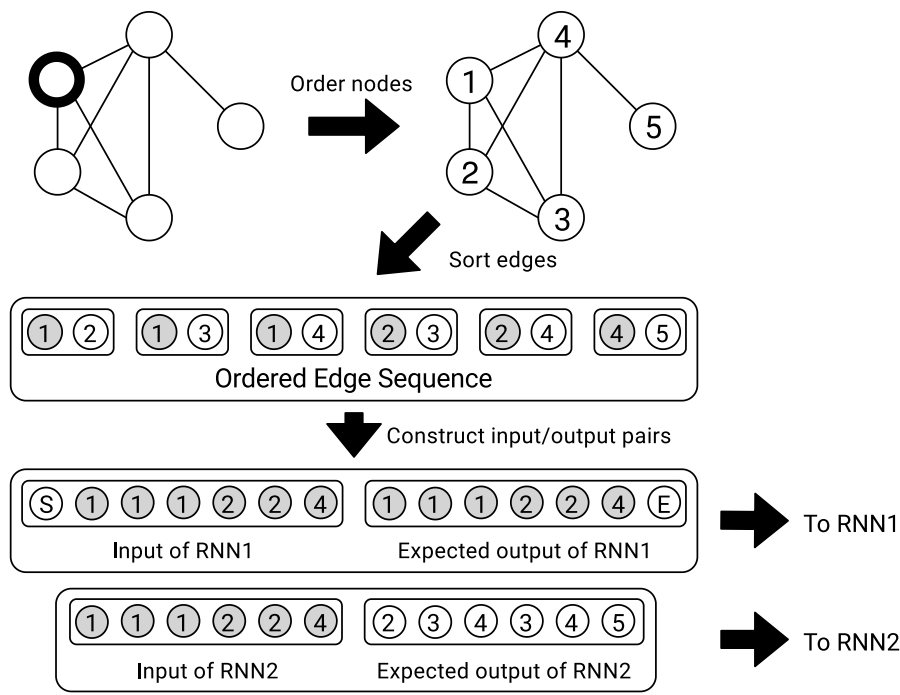}
\caption{An illustration of the preprocessing workflow applied to each graph before training. The node with the thick border in the leftmost graph is the one randomly selected as the starting node for the node ordering procedure. Shading is used to indicate whether a node ID belongs to the source sequence (shaded) or to the destination sequence (not shaded). S and E are special symbols that denote the start and end of the sequence.}
\label{fig:preprocessing}
\end{figure}

Figure \ref{fig:preprocessing} illustrates the preprocessing pipeline applied to each graph before training. First, nodes are ordered by a node ordering procedure $\pi$, and labelled accordingly. Then, edges are sorted in lexicographic order to construct the ordered edge sequence. From that, we extract the source and destination sequences and construct the inputs for the two RNNs. As said previously, RNN1 learns to predict the next symbol of the source sequence, hence receives as input the source sequence without the last node identifier, and expects as output the same node sequence, but shifted one time step ahead. Special Start-Of-Sequence (\texttt{SOS}) and End-Of-Sequence (\texttt{EOS}) symbols are used to mark the beginning and end of the source sequence, respectively. RNN2 learns to predict the endpoint of an edge, given the corresponding source node. As such, it receives as input the source sequence, and expects as output the destination sequence.

A crucial part of the graphs preprocessing is how nodes in the graph are ordered, that is, which function $\pi$ to choose. This is an extremely hard problem in principle, since graphs are invariant to node permutations by definition ($V$ is a set, thus is unordered). Another issue, more subtle, is that even with a consistent ordering, one would require a large quantity of graphs to prevent the model from just memorizing the most frequent generative paths (resulting in generating the same graph over and over), or the generative paths in the training sample (resulting in a model that generates training graphs). While scarcity of data is rarely the case for molecules, were large databases are available, most graph datasets are typically composed of few tens of hundreds of graphs. This makes generative models, especially the most expressive such as the ones based on recurrent neural networks, prone to overfitting and hard to train. 

In this work, we rely on a node ordering procedure inspired by the work in \cite{you34}. The algorithm proceeds as follows: \emph{a}) select one node at random; \emph{b}) use that node as starting point for a breadth-first visit of the graph; \emph{c}) finally, label the nodes according to the breadth-first visit order. In contrast with the original approach, however, we impose that once fixed, the node ordering cannot change during learning. This choice differentiates our method from \cite{you34}, which instead lets the node ordering change at each epoch, choosing a different starting node for the breadth-fist visit at different epochs. Note that their ordering procedure implicitly allows the network to be trained over node orderings induced by every possible node permutation. This acts as a powerful regularizer, since the model learns from potentially very different graphs at each training epoch. Our approach, instead, is to keep the node ordering consistent during the entire learning phase, and prevent overfitting with common regularization techniques such as dropout. In Section \ref{sec:ordering}, we evaluate experimentally our approach against different node ordering procedures, including the original in \cite{you34}, and show how the node ordering of choice is well-suited to make our model generate good quality graphs.

\subsection{Training}
Before training, graphs are transformed into their corresponding ordered edge sequences, from which their related source and destination sequences are extracted. The source sequence $\mathcal{X} = (x_0, x_1, \ldots, x_{|\mathcal{X}|})$, with $x_0 = \texttt{SOS}$ is given as input to RNN1, with all the elements one-hot encoded. The dimension of the one-hot vector is equal to the largest node ID in the dataset plus 2 (for the \texttt{SOS} and \texttt{EOS} tokens). The model computes the probability vector of the next node ID in the sequence $x_i$ as follows:
$$p(x_{i+1}|x_{i}, \mathbf{h}_{i}^p) = \mathrm{SM}(\mathrm{Lin}(\mathrm{GRU}(\mathrm{Emb}(x_{i}), \mathbf{h}_i^p)))),$$
where Emb is an embedding layer where node IDs are first projected, GRU is the recurrent unit that processes the node ID embedding along with the current state $\mathbf{h}_i^p$, Lin is a linear projection to map the recurrent output to node ID space, and SM is a softmax layer which transforms its input into a probability distribution. Note that the GRU layer produces, besides the usual output, the next hidden state $\mathbf{h}_{i+1}^p$ as well (omitted above), which is fed as input to the next time-step along with $x_{i+1}$. The initial hidden state $\mathbf{h}_{0}^p$ is set to be the zero vector. This network is trained to minimize the following categorical cross-entropy:
$$\mathcal{L}_{\mathrm{RNN1}}(\mathcal{X}) = - \frac{1}{|\mathcal{X}|} \sum_{i=0}^{|\mathcal{X}|} x_{i+1} \log p(x_{i+1}\mid x_i, \mathbf{h}_i^p),$$
In other words, RNN1 minimizes the negative log-likelihood between the predicted probability of the next node ID in $\mathcal{X}$ and the expected node ID. Once all of $\mathcal{X}$ is processed, we set $\mathbf{x}_{|\mathcal{X}|}^p = \mathbf{h}_0^q$, i.e. the last recurrent state of RNN1 is used to initialize the state of RNN2, and the process moves to RNN2. The second network is given $\mathcal{X}$ as input, and computes the probability distribution of the endpoint $y_i$ associated to the node ID $x_i$ as follows:
$$q(y_{i}|x_{i}, \mathbf{h}_{i}^q) = \mathrm{SM}(\mathrm{Lin}(\mathrm{GRU}(\mathrm{Emb}(x_{i}), \mathbf{h}_i^p)))),$$
essentially using the same architecture as RNN1. Note that for both RNN1 and RNN2 we use teacher forcing \cite{williams25}, i.e we feed to the network the ground truth node IDs instead of the predicted ones. RNN2 is trained to minimize the following categorical cross-entropy:
$$\mathcal{L}_{\mathrm{RNN2}}(\mathcal{Y}) = - \frac{1}{|\mathcal{Y}|} \sum_{i=0}^{|\mathcal{Y}|} y_i \log q(y_{i}\mid x_i, \mathbf{h}_i^q),$$
where we adopt the same conventions as the first loss function (remember that $|\mathcal{X}| = |\mathcal{Y}|$). In short, RNN2 minimizes the negative log-likelihood between the predicted probability of the node IDs in $\mathcal{Y}$, and the expected node IDs. Note that the pair ($x_i, y_i$) is effectively an edge of the training graph. 

\begin{figure}[ht!]
\centering
\includegraphics[width=.6\textwidth]{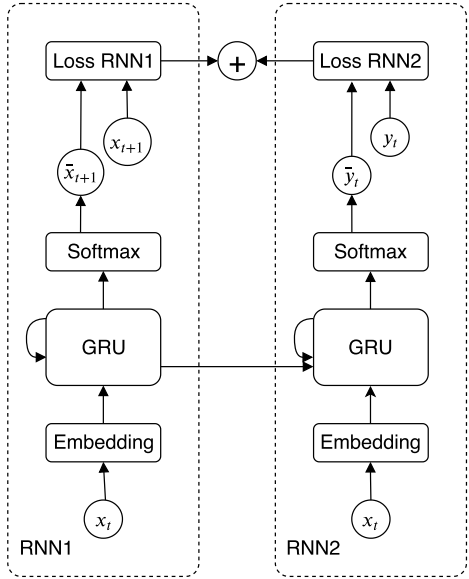}
\caption{A depiction of the feed-forward architecture of the model at training time. Here, circular nodes indicate sequences, and solid line boxes indicate neural network layers. Sequences are hatted if they are network predictions, or not hatted if they are ground truth. We used Gated Recurrent Units (GRU) \cite{cho38} as recurrent layers. In the GRU layer, the self-looping arrow on the left indicates the state that is passed along each time step $t$, while the arrow that connects the GRU layers of the two networks indicates that the last hidden state of the leftmost GRU is used to initialize the rightmost GRU. For ease of visualization, the linear layer that maps the GRU output to node ID space is not shown.}
\label{fig:training}
\end{figure}

The final objective function is obtained by summing out the two losses over all input sequences: 
$$\mathcal{L}\, =\, \frac{1}{|S_\mathcal{G}|} \sum_{(\mathcal{X}, \mathcal{Y}) \in S_{\mathcal{G}}}  \mathcal{L}_{\mathrm{RNN1}}(\mathcal{X}) + \mathcal{L}_{\mathrm{RNN2}}(\mathcal{Y}),$$
and the whole architecture is back-propagated in an end-to-end fashion. During training, we record the overall loss at each epoch, and choose as best model the one with the lowest loss among all training epochs. Figure \ref{fig:training} illustrates the training process.

\subsection{Inference}
\begin{figure}[h!]
\centering
\includegraphics[width=\textwidth]{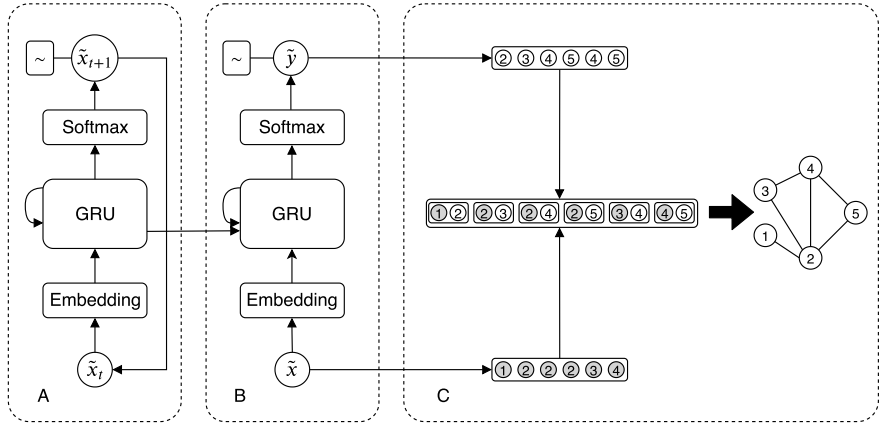}
\caption{An illustration of the inference procedure of the proposed model. Circular nodes represent sequences, while solid line boxes represent neural network layers. In block A, a novel source sequence is sampled from RNN1. The box with the tilde represents a sampling operation. The arrow going from $\tilde{x}_{t+1}$ to $\tilde{x}_t$ indicates that the output at time step $t$ becomes the input for the network at time step $t+1$. In block B, the source sequence sampled from RNN1 is used as input to RNN2, and conditions the generation of the destination sequence $\tilde{y}$. In block C, the two sampled sequences are combined together, generating a novel graph.}
\label{fig:inference}
\end{figure}

Inference starts generating a novel source sequence $\mathcal{X}$ by sampling from $P(\mathbf{X})$, that is, RNN1. More specifically, at each step, a new node ID is sampled from the softmax layer of the network, according to the inferred categorical distribution on node IDs, conditioned on the input node and the current state of the sequence. This sampled node IDs is next as input for the next step. The first node ID of the sequence is obtained by deterministically feeding the \texttt{SOS} symbol as input. The generative process is iterated until the \texttt{EOS} token is encountered. After the novel $\mathcal{X}$ sequence is sampled, it is used as input for RNN2 (whose state $\mathbf{h}$ is initialized with the last recurrent state of RNN1). A novel destination sequence $\mathcal{Y}$ is then generated by sampling from $P(\mathbf{Y}|\mathbf{X})\; P(\mathbf{X})$, i.e. RNN2. More in detail, a node ID is sampled from the inferred categorical distribution of destination nodes $\mathcal{Y}$, conditioned on the input and the state. Generation stops once the \texttt{EOS} token sampled before is encountered as input. After both sequences are obtained, a novel ordered edge sequence can be reconstructed by pairing the two generated sequences. This novel ordered edge sequence completely determines the node and edge sets of a novel graph. Figure \ref{fig:inference} illustrates the inference phase. Note that, since we need to sample each network for a number of steps equal to the length of an ordered edge sequence twice, the inference process has time complexity roughly of $O(2m)$ in the worst case, where $m$ is the maximum number of edges for all graphs in the training sample.

\subsection{Implementation details}\label{sec:implementation}
The model is implemented using the \texttt{PyTorch}\footnote{\texttt{https://github.com/marcopodda/grapher}} \cite{paszke26} library; after an initial exploratory phase (not reported), we found a suitable set of hyper-parameters which we fixed before training. We choose among 32 and 64 for the embedding dimension, 128 and 256 for the recurrent state size, 0.25 and 0.35 of dropout rate. As recurrent cells, we used 2 layers of Gated Recurrent Units (GRU) \cite{cho38}. We used the Adam \cite{kingma36} optimizer with learning rate of 0.001, halved every 200 epochs. We trained the models for a maximum of 2000 epochs, stopping training whenever the loss started to plateau (we found a suitable empirical threshold after running a number of exploratory training instances). As regards the inferential part, we choose temperatures to sample the softmax layers of the two networks among $0.5$, $0.75$ and $1.0$.

\section{Experiments}\label{sec:experiments}
Following, we detail the experiments to assess our model, describing the datasets used for learning, the baselines we compared to, and the framework used for evaluation.

\subsection{Datasets}\label{sec:datasets}
In our experiments, we evaluate our model on 5 datasets. The choice of datasets is driven by the high-level purpose of our model, which, we refrain, is not to approximate a particular graph distribution, but to be able to learn a broad variety of complex nodes and edges dependencies. As such, we choose a set of datasets very different among themselves, as regards graph structure. Concretely, we used the following datasets:

\begin{itemize}
    \item \emph{Ladders}, a synthetic dataset of ladder graphs. We created the dataset generating all possible ladder graphs having number of nodes $2N, N = 2, \ldots, 20$, and replicated them 10 times, resulting in a total size of 180 graphs. This dataset is inspired from the grids dataset employed in \cite{you34}. However, we resorted to ladder graphs because they have the same node degree distribution. For further explanations on the role of the Ladders dataset, see the end of this section;
    \item \emph{Community}, a synthetic dataset of graphs with two-community structure, i.e. graphs composed of two densely connected clusters of nodes, which are weakly connected among themselves. Here, we want to capture both the strong connectivity inside the single cluster, as well as the weaker connectivity between communities. Such dependency is common in social and biological networks (for example, densely connected communities in metabolic networks often represent functional groups; see e.g. \cite{girvan27}). To generate graphs for this dataset, we firstly generate two clique graphs of random size between $8 \leq N \leq 20$. We then remove some edges randomly from the two clusters with probability $0.4$, and then connect the two communities with 1 or 2 edges at random. The generated dataset is composed of 1000 graphs. This dataset shares a similar construction with the community dataset used in \cite{you34}, which we could not employ directly due to lack of reproducibility.
    \item \emph{Ego}, a dataset of ego networks extracted from the Citeseer dataset \cite{citeseer31}. In this case, the dependency to capture is the presence of a focal node (the "ego" node), which is connected to the majority of nodes in the graph, while the other nodes (the "alter" nodes) have weaker connectivity. This dependency is often seen in social networks-like graphs. Note that ego graphs are usually well-suited to be modeled by the BA model. In our experiments, we extracted all possible ego networks of radius 2 (i.e. the maximum path length between the ego node and the alter nodes is 2) from the Citeseer dataset, resulting in a total size of 1719 graphs.
    \item \emph{Enzymes}, a subset of 436 graphs taken from the ENZYMES dataset \cite{enzymes32}.
    \item \emph{Protein}, a subset of 794 graphs taken from the Dobson and Doig \cite{dobson28} dataset.
\end{itemize}
The last two datasets are composed of real-world chemical compounds, represented as graphs where nodes identify atoms, and edges identify chemical bonds between them. In their case, the goal is to capture patterns of connectivity typical of molecules, such as functional groups or aromatic rings.

All datasets have a number of nodes comprised between 4 and 40, and a number of edges comprised between 2 and 130. Before training, we held out a portion of the dataset from testing purposes. In particular, for all datasets except Ladders, we set aside 30\% of graphs, and used them for evaluation purposes. This means that we evaluate our model on graphs that were not seen during training, considering them as a random sample drawn from their corresponding unknown graph distribution. For the Ladders dataset, we adopt a different setup: since the dataset is effectively composed of 18 different graphs replicated 10 times, we held out 10\% of graphs in a stratified fashion, meaning that the held-out set for Ladders is composed of the same 18 ladder graphs found in the training set. To motivate this choice, let us clarify the role of the Ladders dataset: its purpose is not to learn to generalize to unseen ladders, but to show that random models are not able to learn their degree distribution. In fact, ladder graph nodes have an almost constant node degree of 3, except for nodes at the corner, which have degree of 2. This dependency is indeed very unlikely to be learned modelling connectivity in a random fashion. Dataset statistics are presented in Table \ref{tab:datasets}. 

\begin{table}[]
    \footnotesize
    \centering
    \begin{tabular}{ccccc}
        \toprule
         \textbf{Dataset} & \textbf{No. of graphs} & \textbf{Test size} & \textbf{Avg. no. of nodes} & \textbf{Avg. no. of edges}\\
         \midrule
         Ladders & 180 & 18 & 21.00 & 29.50 \\
         Community & 1000 & 300 & 24.38 & 88.58 \\
         Ego & 1719 & 516 & 13.02 & 18.51 \\
         Enzymes & 436 & 130 & 26.14 & 51.16 \\
         Protein & 794 & 238 & 20.49 & 38.67 \\
        \bottomrule
    \end{tabular}
    \caption{Statistics of datasets used in the experiments.}
    \label{tab:datasets}
\end{table}

\subsection{Baselines}
We compare the performances of our model to those of four baseline models. Two of them are classical generative models of graphs coming from graph theory literature, namely the ER and BA models. The rationale behind this choice is to assess whether our model is able to perform better than random models that do not take into account edge dependencies (ER), or consider just one simple edge dependency (BA, where the enforced dependency is the number of nodes in a graph a new node gets attached to). 

The ER model has two parameters: $n$, the number of nodes of the graph to be generated, and $p$, the probability of adding an edge between a pair of nodes. The generative process that they enforce can be described informally as follows: first, pick $n$, then, for each possible pair of nodes, sample a Bernoulli random variable with probability $p$, and connect the two nodes with an edge (or not) based on whether its value exceeds $p$. We optimize these two parameters picking $n$ among the number of nodes of graphs in the training sample, and choosing $p$ from a grid of possible values, such that the earth mover distance between the training graph statistics and the generated graph statistics is minimized.

The BA model has two parameters as well: besides $n$, it has a parameter $m$ that specifies the number of edges to attach from a new node to nodes existing in the current graph. The generative process for a BA graph proceeds incrementally: given an already formed graph, add a new node and connect it (or not) to  a destination node with success probability proportional to its degree. In our experiments, the two parameters of the BA model are optimized in a similar fashion as the ER model.

We compared our model to GRNN (described in Section \ref{sec:related_work}). This choice is mainly motivated by the fact that we wanted to compare the performances of our model against a strong, state-of-the-art competitor. We implemented the model according to the original code repository provided by the authors, following their implementation and their hyper-parameters configuration closely.

Lastly, we introduce a third baseline model, a recurrent neural network with GRU cells that is trained to predict the adjacency matrix of the graph, entry by entry. We will call this baseline GRU-B(aseline) from now on. It is arguably the simplest model one can come up with to model graph generation in a recurrent fashion, with the limitation that it has to sample $N(N-1)/2$ entries to fully specify the adjacency matrix of an undirected graph, making it prone to learning issues induced by long-term dependencies \cite{bengio29}. Another drawback of this model is that, since it is trained on adjacency matrices alone, it can hardly learn higher-order motifs. In this sense, it has limited expressiveness with respect to the proposed approach and GRNN, which are, at least in principle, capable of learning higher-order connectivity patterns.

\subsection{Evaluation Framework}\label{sec:framework}
We assess our model against the competitors following a quantitative and qualitative approach. In a first, quantitative experiment, we evaluate how models are able to keep generating different graphs on large samples. This is a crucial aspect, as generated copies of graphs seen on the training set, as well as multiple copies of the same graphs, are clear signs of overfitting. Moreover, to detect such overfitting, generated samples must be large, as small samples might misleadingly indicate good generalization just as a consequence of a "lucky" draw. Thus, we generate two samples of size 1000 and 5000 from our model and the competitors, and for each of the two samples we measure two scores: \emph{novelty}, defined as the proportion of generated graphs which are not present in the training sample, and \emph{uniqueness}, defined as the proportion of graphs which do not appear more than once in the generated sample. Note that, especially in the Cheminformatics domain, one is also interested in the validity of generated graphs, defined as the proportion of generated graphs that represent a valid molecule under chemical constraints such as valency \cite{liu17}. We, on the other hand, do not take into account validity, for two reasons: first, our model does not consider node and edge labels, which are essential to test for molecular validity; second, validity is not defined in a general sense, but is rather a domain-specific concept. We also measure the time each model takes to generate the 5000 graph sample, to get a sense of how fast models are at generating novel graphs.

In a second experiment, we assess how much samples from the models resemble a random sample from the graph distribution of reference. To do so, we sample from all the models a number of graphs equal in size to the held-out test set, and evaluate the Kullback-Leibler Divergence \cite{kullback35} (KLD) between a set of graph statistics collected both on our samples and on the test sample. 
We remind that the KLD is defined as: 
$$\mathrm{KLD}(P,Q) = - \sum_{x \in X} P(x) \log (Q(x) / P(x)),$$
where $P$ and $Q$ are arbitrary distributions. Note that the KLD is 0 iff $P = Q$, i.e. the two distributions match perfectly (hence, lower values are better). In our case, $P$ is the empirical distribution of some statistics calculated on the test set, $Q$ is the empirical distribution of some statistics calculated on the generated sample, and $X$ is a finite vector containing the computed statistic. In this work, we choose to evaluate:

\begin{itemize}
    \item Average Degree Distribution (ADD), that is, the empirical node degree distribution averaged over all graphs in the sample. Generative models that are able to capture the correct node degree distribution do so by correctly capturing local connectivity patterns.
    \item Average Clustering Coefficient (ACC), that is, the empirical clustering coefficient  distribution averaged over all graphs in the sample. Intuitively, the clustering coefficient of a node specifies of how much it is close to being the node of a 3-clique (or triangle); it is a nearly local metric, calculated as the ratio between the potential number of triangles the node could be part of, and the actual number of triangles the node is part of. It has values in the range $[0,1]$, where $1$ means that the node forms a clique with all its neighbors, while $0$ means that it is the central node of a star subgraph that includes all its neighbors.
    \item Average Orbit Count (AOC), that is, the distribution of counts of node 4-orbits of size averaged over all graphs in the sample. Intuitively, a node $k$-orbit is a generalization of the clustering coefficient to more complex structures than triangles with $k$ nodes, and an orbit count specifies how many of these substructures the node is part of. This measure is useful in order to understand if our model is capable to match higher-order graph statistics, as opposed to node degree and clustering coefficient, which represent measures of local (or close to local) proximity. In our case, for each node in a graph, we collected the orbit counts of 4 different substructures of 4 nodes. 
\end{itemize}

The KLDs of the statistics of interest are computed as follows: first, we calculate the statistic of each node, for each graph in the sample of interest. Then, for each graph, we concatenate all the values into a vector. Since each graph can have a different number of nodes, such vectors have different lengths; hence, we concatenate each of these vector further into a unique vector for the whole sample, and we fit an histogram of 100 bins to its values. Finally, we compute the KLD between the two histograms (note that both histograms obtained on the test and generated samples are the empirical distributions $P$ and $Q$ in the KLD formula, respectively). Calculations are repeated 10 times, each time drawing a novel sample from the models.

\section{Results and discussion}\label{sec:results}

\subsection{Quantitative analysis}\label{sec:quantitative_analysis}
Table \ref{tab:scores_quantity} shows the results of our quantitative experiments. The best performances in terms of novelty and uniqueness are obtained consistently by the ER and BA models; this was expected, since by definition they produce random graphs, hence very likely to be different from the ones in the training set, as well as different with respect to other graphs in the generated sample. One exception is the Ego dataset, where random models score poorly with respect to the competitors. We argue that this result is due to the nature of the Ego dataset, which is composed of graphs with very scarce connectivity (except for the ego nodes) and a very low number of cycles. With such characteristics, it is easier for a random model to produce duplicates or overfit the training sample, just by setting the parameters that regulate connectivity to small values. 

\begin{table}[]
    \footnotesize
    \centering
    \begin{tabular}{lcccccccc}
        \toprule
         \textbf{Dataset} & \textbf{Metric} & \textbf{ER} & \textbf{BA} & \textbf{GRU-B} & \textbf{GRNN} & \textbf{Ours} \\
         \midrule
          & Novelty@1000              & $0.9990$ & $1.0000$ & $0.8130$          & $0.9870$          & $\textbf{0.9990}$\\
          & Novelty@5000              & $0.9984$ & $1.0000$ & $0.8324$          & $0.9868$          & $\textbf{0.9970}$\\
          Ladders & Uniqueness@1000   & $0.9328$ & $0.6830$ & $0.0560$          & $\textbf{0.5760}$ & $0.2380$\\
          & Uniqueness@5000           & $0.8945$ & $0.5062$ & $0.0330$          & $\textbf{0.3908}$ & $0.1910$\\
          & Time@5000                 & $<$1s    & 1s       & 6m34s             & 17m16s            & 3m35s\\
         \midrule
          & Novelty@1000              & $1.0000$ & $1.0000$ & $\textbf{1.0000}$ & $\textbf{1.0000}$ & $\textbf{1.0000}$\\
          & Novelty@5000              & $1.0000$ & $1.0000$ & $\textbf{1.0000}$ & $\textbf{1.0000}$ & $\textbf{1.0000}$\\
          Community & Uniqueness@1000 & $0.9980$ & $1.0000$ & $\textbf{1.0000}$ & $\textbf{1.0000}$ & $\textbf{1.0000}$\\
          & Uniqueness@5000           & $0.9996$ & $1.0000$ & $\textbf{1.0000}$ & $\textbf{1.0000}$ & $0.9998$\\
          & Time@5000                 & 3s       & 5s       & 7m19s             & 52m15s            & 8m49s\\
         \midrule
          & Novelty@1000              & $0.8179$ & $0.8610$ & $0.7540$          & $\textbf{1.0000}$ & $0.9740$\\
          & Novelty@5000              & $0.8166$ & $0.8482$ & $0.7386$          & $\textbf{1.0000}$ & $0.9588$\\
          Ego & Uniqueness@1000       & $0.6369$ & $0.7120$ & $0.5910$          & $\textbf{1.0000}$ & $0.9390$\\
          & Uniqueness@5000           & $0.5206$ & $0.6354$ & $0.4146$          & $\textbf{1.0000}$ & $0.9044$\\
          & Time@5000                 & 1s       & 1s       & 6m06s             & 1h10m14s          & 3m05s\\
         \midrule
          & Novelty@1000              & $1.0000$ & $1.0000$ & $0.9960$          & $0.9890$          & $\textbf{0.9990}$\\
          & Novelty@5000              & $0.9996$ & $1.0000$ & $0.9964$          & $0.9822$          & $\textbf{0.9994}$\\
          Enzymes & Uniqueness@1000   & $0.9940$ & $0.9820$ & $0.7770$          & $\textbf{0.9900}$ & $0.9780$\\
          & Uniqueness@5000           & $0.9900$ & $0.9796$ & $0.5398$          & $\textbf{0.9668}$ & $0.9618$\\
          & Time@5000                 & 4s       & 3s       & 7m16s             & 52m21s            & 4m46s\\
        \midrule
          & Novelty@1000              & $0.9920$ & $1.0000$ & $0.9100$          & $0.8640$          & $\textbf{0.9240}$\\
          & Novelty@5000              & $0.9896$ & $1.0000$ & $0.9144$          & $0.8432$          & $\textbf{0.9306}$\\
          Protein & Uniqueness@1000  & $0.9710$ & $0.9080$ & $0.7330$          & $0.8880$          & $\textbf{0.9010}$\\
          & Uniqueness@5000           & $0.9428$ & $0.8976$ & $0.4816$          & $0.8242$          & $\textbf{0.8718}$\\
          & Time@5000                 & $<$1s    & 4s       & 6m22s             & 54m12s            & 3m38s\\
          \bottomrule
    \end{tabular}
    \caption{Results of the quantitative analysis of the generated samples. In the leftmost column, both the metric of interest as well as the sample size (either 1000 or 5000) is specified. Best performances of models based on RNNs are bolded.}
    \label{tab:scores_quantity}
\end{table}

In contrast, our model and GRNN consistently generate graphs with high novelty and uniqueness rates in almost all scenarios. The only exception to this trend is the ladders dataset, where both our model and GRNN score poorly, while random models score higher. However, we remark that Ladders is an ill-constructed dataset, as mentioned in Section \ref{sec:datasets}. The main purpose of that particular dataset is not to quantify novelty or uniqueness of the samples, but to show that random models are unable to learn the strong edge dependency of ladder graphs. In a sense, the fact that random models score higher provides evidence  that they are unable to reproduce the connectivity patterns of ladder graphs: higher scores are likely due to the fact that they have fit random connectivity parameters to the training sample, resulting in very heterogeneous graphs. For further evidence, we refer the reader to the qualitative analysis, where results show clearly that random models generate low quality ladder graphs.

The GRU-B model greatly under-performs in all but the Community dataset. One could legitimately infer that the model is overfitting and has just memorized training adjacency matrices, instead of more general connectivity paths; however, both from the qualitative analysis and by observing that the training loss (not shown) for the model has plateaued at a high loss, it is safe to affirm that the model cannot perform any better. This provides evidence that a simple recurrent model such as GRU-B, at least in this form, is not well suited for the task of graph generation. 

Our model performs close to a 1.0 rate in novelty and uniqueness in every dataset except Ladders (for reasons explained above), with the notable mention of the Protein dataset, where it obtains the best performances in every scored metric with a margin of 0.02 to 0.06 points with respect to the GRNN model. On the other hand, the GRNN model obtains the best scores in the Ego dataset, scoring 1.0 in every considered metric, beating our model with a margin of 0.3 to 0.1 points. However, our model is the most consistent across all the datsets, performing over 0.87 in all considered scores in every dataset except Ladders.  

Note how all models generate novel and unique graphs in the Community dataset, with a rate of 1.0 or very close. This can be explained by considering the nature of the Community dataset, whose graphs are essentially composed by two random graphs weakly connected among each other. Thus, since generating a graph from that distribution is very similar to generating one at random, samples are highly likely to be different from each other.

As regards sampling time, we note that random models are the fastest during generation; this was expected as well, since they have only 2 parameters to sample from, while all RNN-based models have a larger number of parameters. Among the RNN-based models, our model is the fastest at generating new samples, the only exception being the Community dataset, where however it elapses only 1 minute more than the winning model, GRU-B. In contrast, the GRNN model has sampling times 5 to 20 times slower than our model. For completeness, however, we report that while we sampled from the GRNN model with batch size of 1 to achieve a fair comparison, its implementation allows to draw samples in batches, greatly speeding up sampling time.

Table \ref{tab:mean_rank_quantity} shows the mean rank of the models for each considered metric (except time), averaged over all datasets. More precisely, given one metric, we collect the scores of all models on that particular metric, sort the corresponding vector in descending order from highest to lowest score, and assign as rank the index associated to the position of the models performance in the sorted vector. The ranks are finally averaged over all datasets, to provide a measure of how models behave globally. Results show that the BA model has the highest mean rank as regards novelty on a sample size of 1000 (for the reasons discussed above), while GRNN has the highest mean rank as regards novelty on a sample of 5000. Our model obtains the highest mean rank on uniqueness, on both sample sizes, and also on novelty for a sample size of 1000 among RNN-based models.

\begin{table}
    \footnotesize
    \centering
    \begin{tabular}{cccccccc}
          \toprule
          \textbf{Metric} & \textbf{ER} & \textbf{BA} & \textbf{GRU-B} & \textbf{GRNN} & \textbf{Ours} \\
          \midrule
          Novelty@1000    & $3.0000$ & $2.6000$ & $3.2000$ & $3.2000$          & $\textbf{3.0000}$\\
          Novelty@5000    & $3.0000$ & $2.6000$          & $4.0000$ & $\textbf{2.4000}$ & $3.0000$\\
          Uniqueness@1000 & $2.6000$ & $3.4000$          & $3.2000$ & $3.4000$          & $\textbf{2.4000}$\\
          Uniqueness@5000 & $2.6000$ & $4.0000$          & $3.4000$ & $2.8000$          & $\textbf{2.2000}$\\
          \bottomrule
    \end{tabular}
    \caption{Mean ranks on all considered quantitative metrics except time, obtained by the examined models over all datasets. Best performances of models based on RNNs are bolded.}
    \label{tab:mean_rank_quantity}
\end{table}

\subsection{Qualitative evaluation}\label{sec:qualitative_analysis}
Table \ref{tab:results_quality} shows the results of the qualitative evaluation of the models. 
It can be clearly seen how random models are not able to learn complex dependencies, scoring poorly on all datasets in every considered metric, in contrast with the RNN-based models. One exception is the Ego dataset, where both the ER and BA model obtain the best KLD for the AOC metric (orbit counts) with 0.07 and 0.09, respectively (standard deviation intervals overlap). This good result is in accordance with the argument exposed in Section \ref{sec:quantitative_analysis}, where we explained that random models have likely fitted the dataset with very small values of the parameters that regulate connectivity. Indeed, ego graphs have usually a very low higher-order connectivity, hence generating graphs with weak connectivity results in capturing the correct orbit counts distribution. As regards the Ladders dataset, the BA model obtains the best ACC (clustering coefficient), scoring $0$ in all the repeated experiments. Again, a similar argument can be formulated to explain such good performance: the clustering coefficient of ladder graph nodes, which is 0 by construction, can be easily captured by random models by generating very weakly connected graphs. However, note how the BA model completely fails at generating graphs with the correct node degree distribution (most likely generating graphs with degree of 1 in most nodes), scoring the worst among all datasets. The same argument applies to the ER model, providing evidence that random models are only able to capture a limited set of graph properties, sacrificing many others.

\begin{table}[]
    \tiny
    
    \noindent
    \makebox[\textwidth]{
    \begin{tabular}{lcccccc} 
        \toprule
         \textbf{Dataset} & \textbf{Metric} & \textbf{ER} & \textbf{BA} & \textbf{GRU-B} & \textbf{GRNN} & \textbf{Ours} \\
         \midrule
          & ADD          & $0.6338 \pm 0.0604$ & $1.0091 \pm 0.1534$ & $0.6266 \pm 0.1348$ & $\textbf{0.0013} \pm 0.0014$ & $\textbf{0.0100} \pm 0.0038$\\
         Ladders & ACC   & $0.2146 \pm 0.0569$ & $\textbf{0.0000} \pm 0.0000$ & $\textbf{0.0047} \pm 0.0058$ & $\textbf{0.0004} \pm 0.0013$ & $\textbf{0.0050} \pm 0.0055$\\
          & AOC          & $1.6927 \pm 0.4817$ & $1.3893 \pm 0.3748$ & $1.7828 \pm 0.3068$ & $0.7562 \pm 0.0534$ & $\textbf{0.3011} \pm 0.1001$\\
         \midrule 
          & ADD          & $1.7754 \pm 0.0790$ & $0.4632 \pm 0.0169$ & $0.1548 \pm 0.0202$ & $0.0892 \pm 0.0065$ & $\textbf{0.0175} \pm 0.0034$\\
         Community & ACC & $1.1399 \pm 0.2115$ & $0.2612 \pm 0.0147$ & $0.1641 \pm 0.0217$ & $\textbf{0.0988} \pm 0.0081$ & $\textbf{0.0927} \pm 0.0201$\\
          & AOC          & $3.4481 \pm 1.2678$ & $2.2806 \pm 0.1714$ & $2.5674 \pm 0.0710$ & $\textbf{0.0758} \pm 0.0057$ & $\textbf{0.0537} \pm 0.0171$\\
         \midrule 
          & ADD          & $0.2545 \pm 0.0162$ & $0.6953 \pm 0.0363$ & $0.3189 \pm 0.0058$ & $0.1205 \pm 0.0392$ & $\textbf{0.0224} \pm 0.0048$\\
         Ego & ACC       & $0.7326 \pm 0.1137$ & $0.5224 \pm 0.0573$ & $0.9007 \pm 0.1965$ & $0.3313 \pm 0.0515$ & $\textbf{0.0657} \pm 0.0096$\\
          & AOC          & $\textbf{0.0782} \pm 0.0103$ & $\textbf{0.0943} \pm 0.0070$ & $1.1800 \pm 0.0726$ & $0.4237 \pm 0.0783$ & $0.1142 \pm 0.0096$\\
         \midrule 
          & ADD          & $3.1524 \pm 0.1549$ & $1.9882 \pm 0.0863$ & $0.5226 \pm 0.0348$ & $\textbf{0.0160} \pm 0.0054$ & $\textbf{0.0172} \pm 0.0037$\\
         Enzymes & ACC   & $2.1973 \pm 0.2652$ & $1.2295 \pm 0.0240$ & $0.5883 \pm 0.1949$ & $\textbf{0.0553} \pm 0.0123$ & $\textbf{0.0542} \pm 0.0276$\\
          & AOC          & $3.9728 \pm 0.9711$ & $4.0739 \pm 0.2896$ & $0.9639 \pm 0.1407$ & $\textbf{0.0518} \pm 0.0119$ & $0.1060 \pm 0.0215$\\
         \midrule 
          & ADD          & $2.3075 \pm 0.1203$ & $1.9096 \pm 0.0583$ & $0.4846 \pm 0.0377$ & $\textbf{0.0379} \pm 0.0040$ & $0.0591 \pm 0.0078$\\
         Protein & ACC  & $1.6133 \pm 0.0836$ & $1.1832 \pm 0.0404$ & $0.3536 \pm 0.0452$ & $\textbf{0.0769} \pm 0.0136$ & $\textbf{0.0640} \pm 0.0175$\\
          & AOC          & $2.6382 \pm 0.1303$ & $2.5981 \pm 0.0976$ & $0.6202 \pm 0.0830$ & $0.1014 \pm 0.0181$ & $\textbf{0.0485} \pm 0.0080$\\
         \bottomrule

    \end{tabular}}
    \caption{Results of the qualitative analysis of the generated samples. The three metrics considered are KLDs calculated on Average Degree Distribution (ADD), Average Clustering Coefficient (ACC), and Average Orbit Count (AOD). Best performances are bolded.}
    \label{tab:results_quality}
\end{table}

Results also highlight how the GRU-B model is not able to generate good quality graphs, and especially fails at capturing orbit count distributions: this poor performance can be motivated by the fact that higher-order dependencies are hardly learnable using the adjacency matrix alone. Note that this performance confirms the argument stated in the quantitative analysis, where we affirm that poor results on quantitative metrics are not a consequence of overfitting (otherwise the model would have scored higher simply by replicating graphs found in the training set), but of a limited expressiveness instead.

Among all models, GRNN and ours perform consistently at the state of the art as regards the quality of generated graphs. In most datasets, they perform indistinguishably, meaning that the interval spanned by the standard deviations on the metric overlap. In one case, namely the Ego dataset, our model outperforms GRNN in all considered metrics. However, for fairness, we remind that results on the Ego dataset are probably slightly biased in favor of our model with respect to GRNN, because samples drawn from our model have a lower uniqueness and novelty rate, as per the quantitative analysis. As such, performances might be biased from repetitions of "good" graphs. While we did not investigate this issue any further, we also note that the margin by which our model outperforms GRNN is the highest among all datasets, with GRNN trailing by margins from 0.1 to 0.3 points. In contrast, note that whenever our model performs worse than GRNN, margins are instead noticeably narrower (the largest being 0.05 in the Enzymes dataset on the AOC metric).

Global performances of the models across all datasets are summarized in Table \ref{tab:mean_rank_quality}, where the mean rank of all models is computed in a similar fashion as described in Section \ref{sec:quantitative_analysis}. Our model records the highest mean ranking in all considered qualitative metrics, confirming once again that our approach is able to approximate structural feature distributions of very heterogeneous graphs with high accuracy.

\begin{table}
    \footnotesize
    \centering
    \begin{tabular}{cccccccc}
          \toprule
          \textbf{Metric} & \textbf{ER} & \textbf{BA} & \textbf{GRU-B} & \textbf{GRNN} & \textbf{Ours} \\
          \midrule
          ADD    & $4.4000$ & $4.6000$ & $2.6000$ & $2.0000$ & $\textbf{1.4000}$\\
          ACC    & $4.4000$ & $4.0000$ & $2.8000$ & $2.4000$ & $\textbf{1.4000}$\\
          AOC    & $4.0000$ & $3.8000$ & $3.0000$ & $2.2000$ & $\textbf{2.0000}$\\
          \bottomrule
    \end{tabular}
    \caption{Mean ranks obtained on the evaluated qualitative metrics by the examined models over all datasets. Best performances are bolded.}
    \label{tab:mean_rank_quality}
\end{table}

\begin{figure}[h!]
\centering
\includegraphics[width=\textwidth]{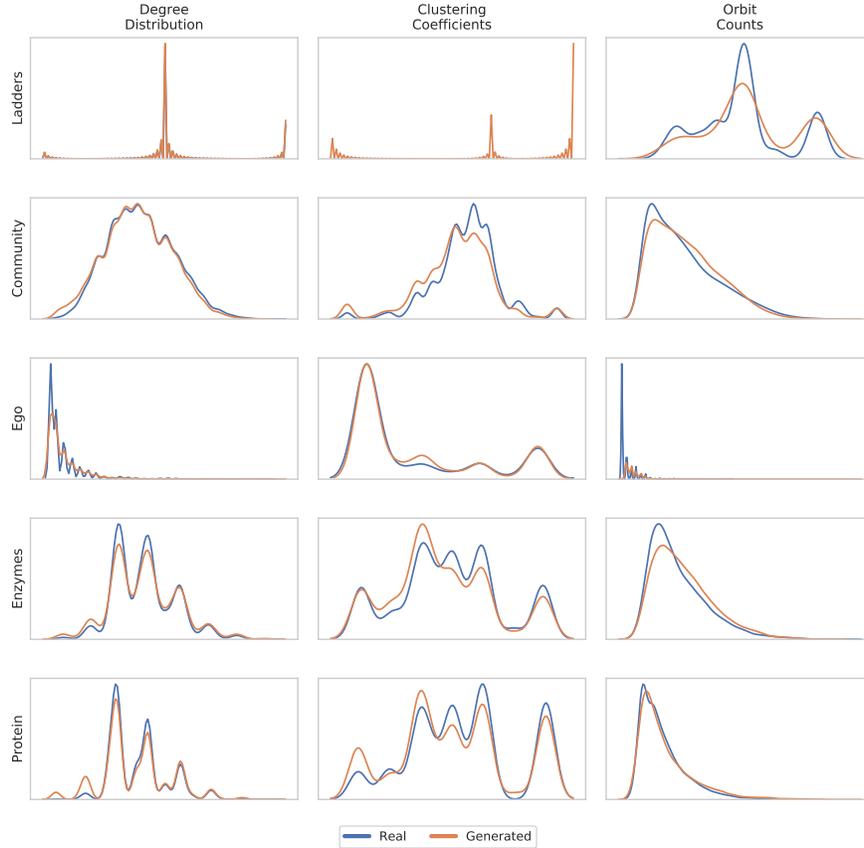}
\caption{Plot of the three qualitative statistics of test samples compared to samples generated by our model. Metrics are displayed by column, dataset by rows. The distribution of the test sample is shown in blue, while the distribution of samples drawn from our model is shown in orange. Scales are omitted since they are normalized, hence not informative.}
\label{fig:distributions}
\end{figure}

To strengthen this concept, in Figure \ref{fig:distributions} we plot the empirical distribution of the considered metrics obtained on the test sample, versus samples generated by our model in every considered dataset. The plots show that, even in cases where the test empirical distribution is skewed or multimodal, the distribution of structural features of the generated samples matches with good precision all the relevant test distribution features, such as long tails or peaks.

\begin{figure}[h!]
\centering
\includegraphics[width=\textwidth]{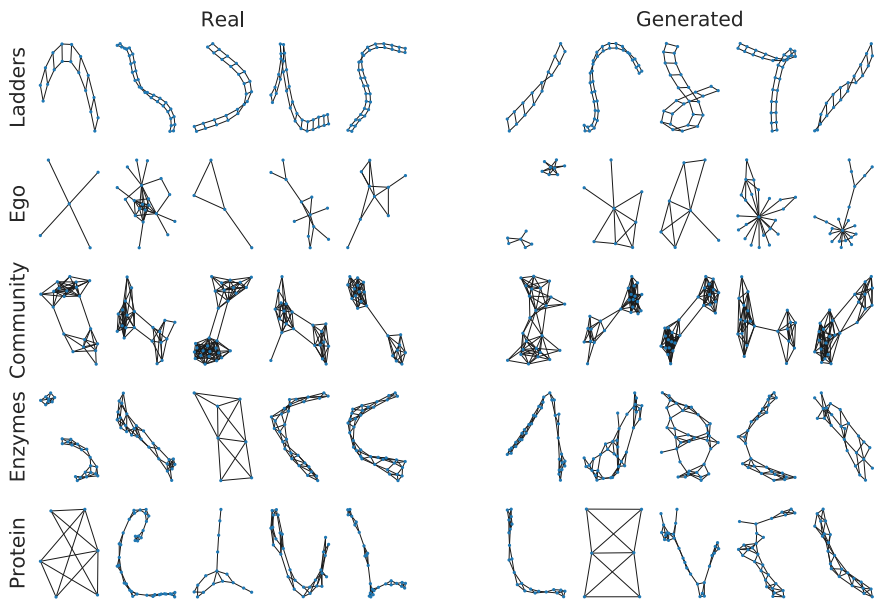}
\caption{Five randomly chosen graphs, both from the training sample (left) and generated by our model (right), on each considered dataset (displayed by row). Notice how our generative model has picked up all the characterizing connectivity patterns.}
\label{fig:samples}
\end{figure}

Finally, in Figure \ref{fig:samples} we show graphs drawn from our model against real graphs drawn from three out of five datasets. Indeed, visual inspection confirms that generated graphs display connectivity patterns detectable in real graphs, for example ego nodes in ego graphs, cliques in protein graphs, and densely connected clusters in the Community dataset.
 
\section{Effect of node ordering}\label{sec:ordering}
As we mentioned earlier, our model is strongly dependent on the type of ordering that is superimposed on graph nodes before training. To this end, we compared our model with 5 variants trained with different ordering strategies. Choices include:

\begin{itemize}
    \item a complete random strategy, which consist in randomly shuffling the node IDs before training, for each graph in the training set. This strategy is expected to perform very poorly, since the model does not see consistent connectivity patterns during training;
    \item the node ordering strategy proposed in \cite{you34}, which we term BF random. Such strategy is very similar to ours (in fact, our strategy is inspired by this), but differs in the fact that the starting node for the breadth-first visit of the graph is selected at random at each epoch. This in principle means that the model is trained on every possible node permutation induced by a breadth-first visit of the graph. On one hand this strategy acts as a regularizer, since at each epoch the training set changes completely. On the other hand, however, changing the node ordering at each epoch could in principle prevent our model from focusing on useful connectivity patterns, simply because they are "masked" away by different node permutations;
    \item a variant of the above mentioned BF random strategy which uses Depth-First Search (DFS) instead of BFS;
    \item only for the Enzymes and Protein datasets, a node ordering strategy induced by the SMILES \cite{weininger37} encoding of the corresponding molecular graph; that is, the order of nodes is given by the position of the corresponding atom label in the SMILES string associated to the molecular graph. Notice that, even though this strategy can in principle produce canonical orderings, it is directly applicable only to molecular graphs who are annotated with atom (node) and bond (edge) labels;
    \item a variant of our proposed strategy which based on DFS.
\end{itemize}

\begin{figure}[h!]
\centering
\includegraphics[width=\textwidth]{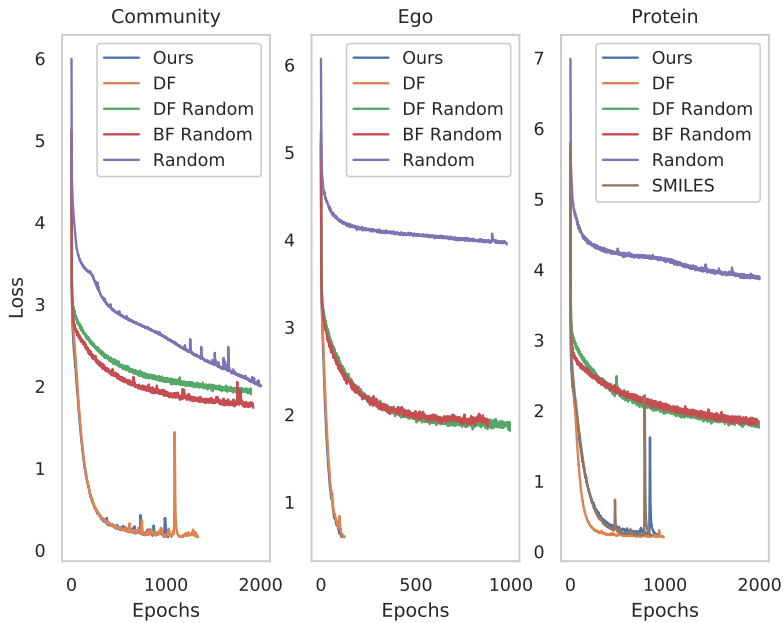}
\caption{Plot of the loss on dataset Protein of variants of our approach trained under different node orderings. Notice how the models trained with our proposed node ordering (blue), and SMILES (orange) are able to reach a lower training loss in a smaller number of epochs, compared to variants trained with random ordering (red) and BF random ordering (green).}
\label{fig:loss_ordering}
\end{figure}

Our first experiment consists in training the models with the alternative node ordering strategies with the same architecture as the proposed model, but trained for a large number of epochs without dropout; in short, trying to overfit the dataset on purpose. This choice is motivated by the fact that we wanted to assess whether the alternative node ordering strategies are able to memorize connectivity patterns (hence, given proper regularization, are able to generalize to unseen connectivity patterns). In Figure \ref{fig:loss_ordering}, we plot the loss obtained by the models, for each considered node ordering strategy, for the Protein dataset. The figure shows that the models trained with our strategy, the variant of our strategy based on DFS, and the SMILES ordering are able to reach a lower loss than the alternative node ordering strategies, which in contrast plateau at higher loss values. This provides evidence that our choice of node ordering strategy is well suited for our model, being sufficiently general to learn connectivity patterns from very different graph datasets. Note that the good results of the SMILES ordering was expected since our node ordering procedure resembles, although is not completely identical, to the way atoms (nodes) are ordered when deriving the SMILES string associated to the molecular graph. However, we also remark that such ordering cannot be applied in general, but it is only suitable for chemical data.

We also conducted a qualitative evaluation similar to the one presented in Section \ref{sec:qualitative_analysis}, where we reused the results of our model and compared them to scores obtained on samples drawn from the three studied variants. Results are detailed in Table \ref{tab:results_ordering}, clearly showing that the model trained with our node ordering strategy is found to be superior in every chosen metric with respect to all assessed variants. The only competitive node ordering strategy is DF, which basically differs from ours in the way graph nodes are visited (DF instead of BF). However, despite being able to match the performances of our node ordering strategy in many cases, the table highlights that it may have problems in approximating structural characteristics in some dataset instances, for example the clustering coefficient in the Community dataset and the orbit count in the Ego dataset. This might be a consequence of the nature of DFS, which does not focus on local patterns first as BF does, but instead explores the graph going through sequential paths, followed by backtracking to the starting node, an approach that is probably harder to train on such datasets. In conclusion, these results further extend the evidences supporting the robustness of our approach. 

\begin{table}
    \tiny
    \noindent
    \makebox[\textwidth]{
    \begin{tabular}{lccccccc}
        \toprule
         \textbf{Dataset} & \textbf{Metric} & \textbf{Random} & \textbf{BF Random} & \textbf{DF Random} & \textbf{DF} & \textbf{SMILES} & \textbf{Ours (BF)}\\
         \midrule
          & ADD           & $0.6427 \pm 0.0941$ & $0.0459 \pm 0.0079$ & $0.0931 \pm 0.0119$ & $0.0590 \pm 0.0060$ & --                           & $\textbf{0.0100} \pm 0.0038$\\
          Ladders & ACC   & $0.3402 \pm 0.0922$ & $0.0306 \pm 0.0117$ & $0.0364 \pm 0.0098$ & $\textbf{0.0058} \pm 0.0056$ & --                           & $\textbf{0.0050} \pm 0.0055$\\
          & AOC           & $2.4868 \pm 0.4030$ & $0.7015 \pm 0.1510$ & $1.1255 \pm 0.1451$ & $0.8513 \pm 0.1602$ & --                           & $\textbf{0.3011} \pm 0.1001$\\
         \midrule
          & ADD           & $0.0382 \pm 0.0043$ & $0.0409 \pm 0.0037$ & $0.0538 \pm 0.0050$ & $0.0445 \pm 0.0060$ & --                           & $\textbf{0.0175} \pm 0.0034$\\
          Community & ACC & $1.0894 \pm 0.1034$ & $0.6144 \pm 0.1304$ & $2.4965 \pm 0.4034$ & $0.5326 \pm 0.0506$ & --                           & $\textbf{0.0927} \pm 0.0201$\\
          & AOC           & $0.4035 \pm 0.0264$ & $0.1022 \pm 0.0086$ & $0.1865 \pm 0.0190$ & $\textbf{0.0544} \pm 0.0077$ & --                           & $\textbf{0.0537} \pm 0.0171$\\
          \midrule
          & ADD           & $0.2720 \pm 0.0087$ & $0.1142 \pm 0.0087$ & $0.0761 \pm 0.0078$ & $0.0594 \pm 0.0127$ & --                           & $\textbf{0.0224} \pm 0.0048$\\
          Ego & ACC       & $0.5057 \pm 0.0338$ & $0.2830 \pm 0.0183$ & $0.2190 \pm 0.0491$ & $\textbf{0.0780} \pm 0.0131$ & --                           & $\textbf{0.0657} \pm 0.0096$\\
          & AOC           & $0.6662 \pm 0.0302$ & $0.6159 \pm 0.0355$ & $0.4277 \pm 0.0220$ & $0.2839 \pm 0.0217$ & --                           & $\textbf{0.1142} \pm 0.0096$\\
         \midrule 
          & ADD           & $0.1179 \pm 0.0093$ & $0.1028 \pm 0.0068$ & $0.1290 \pm 0.0064$ & $0.0282 \pm 0.0033$ & $0.0479 \pm 0.0083$          & $\textbf{0.0172} \pm 0.0037$\\
          Enzymes & ACC   & $1.0667 \pm 0.2747$ & $0.4412 \pm 0.0427$ & $0.9680 \pm 0.0831$ & $\textbf{0.0636} \pm 0.0139$ & $0.1218 \pm 0.0222$          & $\textbf{0.0542} \pm 0.0276$\\
          & AOC           & $0.5498 \pm 0.0530$ & $0.2096 \pm 0.0158$ & $0.4378 \pm 0.0261$ & $\textbf{0.0827} \pm 0.0179$ & $\textbf{0.1343} \pm 0.0214$ & $\textbf{0.1060} \pm 0.0215$\\
         \midrule
          & ADD           & $0.2195 \pm 0.0092$ & $0.1759 \pm 0.0073$ & $0.1845 \pm 0.0059$ & $0.0994 \pm 0.0111$ & $0.1434 \pm 0.0091$          & $\textbf{0.0591} \pm 0.0078$\\
          Protein & ACC  & $0.8169 \pm 0.0578$ & $0.6703 \pm 0.0421$ & $0.9255 \pm 0.0565$ & $0.1712 \pm 0.0198$ & $0.4016 \pm 0.0405$          & $\textbf{0.0640} \pm 0.0175$\\
          & AOC           & $0.5936 \pm 0.0450$ & $0.2021 \pm 0.0136$ & $0.3186 \pm 0.0153$ & $0.0688 \pm 0.0103$ & $0.2171 \pm 0.0268$          & $\textbf{0.0485} \pm 0.0080$\\
         \bottomrule

    \end{tabular}}
    
    \caption{Results of the effect of node ordering on the performances of our model. The variants considered are random ordering, Breadth-First (BF) random ordering as proposed in \cite{you34}, SMILES ordering (only for molecular datasets), ordering of the proposed approach. Best performances are bolded.}
    \label{tab:results_ordering}
    
\end{table}

\section{Conclusions}\label{sec:conclusions}
In this work, we have presented a novel generative model for graphs based on Recurrent Neural Networks, which is capable of generating high quality graphs from very different graph distributions. The key idea at the root of our work is to move from the domain of graphs to the one of sequences to simplify the learning task, while retaining most of the expressiveness. Our motivation to frame the generative process as learning problem on sequences is three-fold: (i) it is intuitive, (ii) it allows to work indirectly on smaller and informative portions of the exponentially large graph space, and (iii) it enables the use of the reliable and heavily optimized machinery of Recurrent Neural Networks to learn effectively. 

With these ideas in mind, we developed a model which, first, orders nodes in a graph sequentially, then converts graphs into ordered sequences of edges, and finally learns these sequences using two RNNs. The first one predicts a sequence of source nodes, while the second uses the output of the first to predict all necessary endpoints to construct a novel graph. We tested our model against canonical random models from graph theory, a RNN baseline with limited expressiveness, and a state-of-the-art model on five different datasets. The evaluation was conducted on multiple levels, considering both quantitative performance measures such as novelty and uniqueness of generated graphs, as well as qualitative metrics to assess whether generated samples retain structural features of those in the training set. The results clearly show that our model performs at, and sometimes surpasses, the state of the art in the task. 
We also conducted a study of how much the procedure chosen to superimpose an order on graph nodes is effective for the particular task of graph generation. More in detail, we compared with 5 variants of our model that were trained on different node orderings. Results show that with the ordering procedure of choice, the model can reach a lower training error, which is essential to learn the complex patterns needed to generate good quality graphs that resemble real-world ones.

The proposed model has also limitations. Even if it works empirically, the constraint imposed by the reliance on a specific ordering is unsatisfactory from a theoretical point of view. For this reason, one key step in future research is to study the node ordering problem more thoroughly, in order to relax and perhaps remove completely such constraint. Another limitation of our approach in its current form is the fact that it is not formulated to include node and edge labels, which are usually found in large classes of graph datasets such as molecular ones. While on one hand this limits the applicability of our approach on those domains, on the other we are also confident that extending the model to include node and edge labels might lead to improvements in generative tasks that deal with those kinds of graphs. Our belief is supported by the intuition that the ability of our model to well approximate structural features of graph distributions, coupled with a reliable mechanism to generate labels, could in perspective improve generalization, since conditioning the generative process not only on structure, but also on features, could help the model learn general connectivity patterns more reliably. This would also allow to compare our model to current molecular graph generator, and investigate the influence of domain specificity on this task.

As for other research directions, we mention the possibility to extend the model by adopting attention mechanisms \cite{bahdanau30}, instead of conditioning the second network only on the last hidden state of the first. Another easily implementable extension of our approach is to model the generative process using a Variational Auto-Encoder in between the two recurrent neural networks, in order to constrain the latent space where the encoding of the first sequence is mapped to be normally distributed. This would have the advantage of cutting sampling time by approximately half, since the inference phase would only require to sample a point from latent space, and decode it using the second network. Lastly, a challenge for the near future is to test whether our approach is capable to scale to larger graphs, which would strengthen our positive results further, as well as expand its applicability to a broader class of problems.

\section*{Acknowledgements}
This work has been partially supported by the Italian Ministry of Education, University, and Research (MIUR) under project SIR 2014 LIST-IT (grant n. RBSI14STDE).

\section*{References}

\bibliography{bibliography}

\end{document}